\documentclass[runningheads]{llncs}
\usepackage[T1]{fontenc}
\usepackage{graphicx}
\usepackage[numbers]{natbib}
\usepackage[utf8]{inputenc} 
\usepackage[T1]{fontenc}    
\usepackage{hyperref}       
\usepackage{url}            
\usepackage{booktabs}       
\usepackage{amsfonts}       
\usepackage{nicefrac}       
\usepackage{microtype}      
\usepackage{xcolor}         
\usepackage{adjustbox}
\usepackage{amsmath}
\usepackage{multirow}
\usepackage[linesnumbered,ruled,vlined]{algorithm2e}
\usepackage{float}
\restylefloat{table}
\usepackage{dblfloatfix}
\usepackage[misc]{ifsym}
\def\id{\mathcal{D}_{\text{ID}}}
\def\idlabel{\mathcal{Y}_{\text{ID}}}
\def\iddata{\mathcal{X}_{\text{ID}}}

\def\ood{\mathcal{D}_{\text{OOD}}}
\def\oodlabel{\mathcal{Y}_{\text{OOD}}}

\begin{document}
%
\title{Revisiting Likelihood-Based Out-of-Distribution Detection by Modeling Representations}
%
\titlerunning{Revisiting Likelihood-Based OOD Detection by Modeling Representations}
%
\author{Yifan Ding\inst{1} \and
 Arturas Aleksandraus\inst{1} \and
 Amirhossein Ahmadian\inst{2} \and
 Jonas Unger\inst{1} \and 
 Fredrik Lindsten\inst{2} \and
 Gabriel Eilertsen\inst{1}}
\authorrunning{Ding et al.}
%
\institute{Department of Science and Technology, Linköping University, Norrköping, Sweden\and Department of Computer and Information Science, Linköping University, Linköping, Sweden\\ {\tt\small firstname.lastname@liu.se}}
\maketitle              
\begin{abstract}
Out-of-distribution (OOD) detection is critical for ensuring the reliability of deep learning systems, particularly in safety-critical applications. Likelihood-based deep generative models have historically faced criticism for their unsatisfactory performance in OOD detection, often assigning higher likelihood to OOD data than in-distribution samples when applied to image data. In this work, we demonstrate that likelihood is not inherently flawed. Rather, several properties in the images space prohibit likelihood as a valid detection score. Given a sufficiently good likelihood estimator, specifically using the probability flow formulation of a diffusion model, we show that likelihood-based methods can still perform on par with state-of-the-art methods when applied in the representation space of pre-trained encoders. The code of our work can be found at \href{https://github.com/limchaos/Likelihood-OOD.git}{\texttt{https://github.com/limchaos/Likelihood-OOD.git}}.

\end{abstract}
\section{Introduction}

Out-of-distribution (OOD) detection is the process of detecting if individual data points, e.g. images, belong to the distribution of training data or not. In machine learning, it is typically used to identify ``unseen'' data points that may lead to unreliable inference. The importance of OOD detection is highlighted by the fact that deep neural networks often perform poorly on input data not part of the training distribution. Detection of OOD input is thus a critical functionality to increase safety and robustness in deployed systems, particularly in real-world applications where the misclassification of OOD samples can lead to severe consequences.  

\begin{figure*}[t]
\centering
\includegraphics[width=\textwidth,trim={0 2.5cm 0 1.5cm},clip]{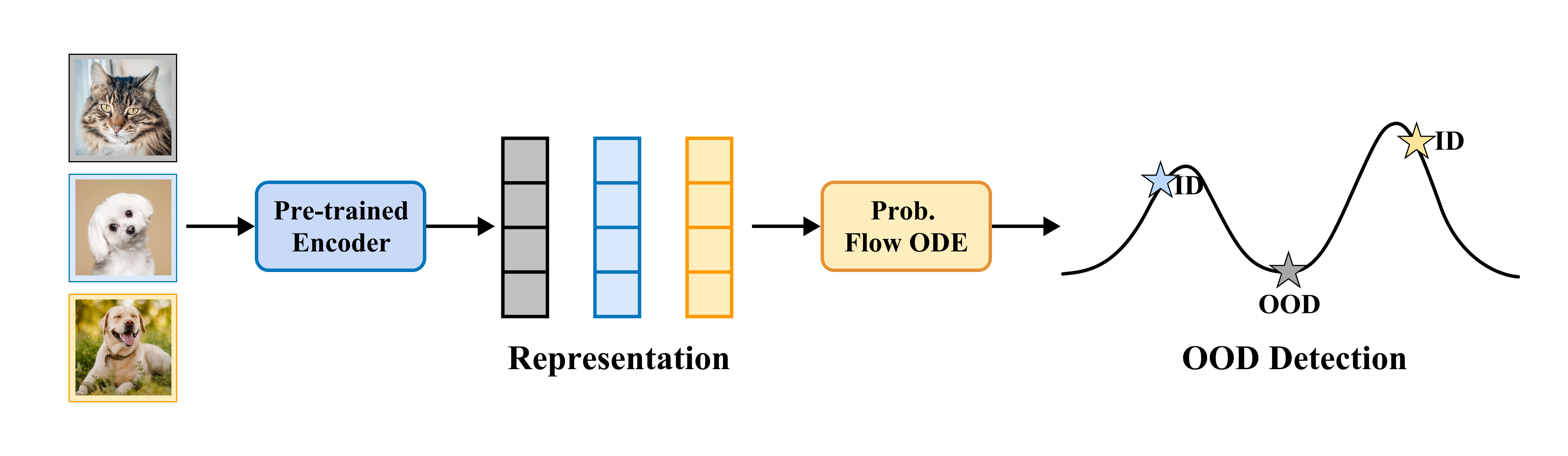}
\caption{Revisiting likelihood-based OOD detection, we employ a score-based diffusion model to detect OOD images in a semantically clustered representation space. The input images are first projected to the representation space. Then, a probability flow ODE \citep{maoutsa2020interacting}, from the corresponding representation score-based diffusion model \citep{song2020score}, is leveraged to calculate the precise likelihood in the representation space.}
\label{fig:RDMOOD}
\end{figure*}
OOD detection was initially developed to identify unseen data with low likelihood given training data and a statistical model \citep{bishop1994novelty}. Explicit likelihood-based deep generative models (DGMs), e.g., autoregressive models and normalizing flows, appear to be well-suited for this purpose. However, it has been demonstrated that these models may assign higher likelihood to OOD images than images from the training distribution \citep{nalisnick2018deep}, e.g., models trained on FashionMNIST and CIFAR-10 incorrectly assign high likelihoods to OOD datasets like MNIST and SVHN where they should assign low likelihood. Meanwhile, it is not obvious how likelihood-based OOD detection methods could generalize to more real world settings (large-scale benchmarks). Due to the problems with likelihood-based OOD detection, the OOD community has mostly shifted the focus to post-hoc methods \citep{hendrycks2019scaling, wang2022vim, sun2022out} based on classifiers. However, these methods usually require supervised training and label information of the in-distribution (ID) data. 
Particularly in the context of medical imaging, such as histopathology data, or other fine-grained categories, obtaining annotations is often expensive or infeasible \citep{pocevivciute2025out} due to the requirement for expert knowledge and regulatory constraints. In such cases, many OOD detection methods that rely on logits from a supervised classifier \citep{msp17iclr, energyood20nips, odin18iclr, hendrycks2019scaling, wang2022vim, liu2023gen} become inapplicable, necessitating alternative approaches that do not depend on labeled training data.

In this paper, we reinvestigate likelihood-based OOD detection and demonstrate that estimating likelihood in the \textit{representation spaces of modern pre-trained image encoders} can be a very promising method for OOD detection.
To this end, we make use of a score-based diffusion model \citep{song2020score} trained on encoded image representations as likelihood estimator.
We argue that the previous failures in \citep{nalisnick2018deep, kirichenko2020normalizing} can be associated with the fact that likelihood estimation carried out in image space is heavily influenced by background statistics \citep{ren2019likelihood} and low-level features \citep{kirichenko2020normalizing}. For example, images with different semantic content may appear close in Euclidean space, while semantically similar images can be far apart due to transformations such as translation. As a result, images tend to cluster based on low-level visual features such as color, luminance, and texture, making it challenging to capture and learn meaningful semantic shifts. However, likelihood works well in semantically clustered representation spaces with accurate likelihood estimation provided by a diffusion model. Unlike \citep{msp17iclr,energyood20nips, hendrycks2019scaling, wang2022vim}, likelihood can work with self-supervised encoders, but it can also leverage ID class labels to guide the diffusion model. As illustrated in Fig.~\ref{fig:RDMOOD}, we first encode images to a representation space. A score-based diffusion model is then trained only on ID representations, and likelihood is estimated for both ID and OOD data at test time. We show that the diffusion model assigns high likelihood to the ID representations and low likelihood to OOD representations. 
Our contributions can be summarized as follows:

\begin{itemize}
    \item We revisit likelihood-based OOD detection by leveraging score-based diffusion models within the representation space of pre-trained image encoders.
    
    \item We conduct extensive experiments on large-scale datasets, evaluating both supervised and unsupervised encoders and benchmarking likelihood against other lines of OOD detection methods.

\end{itemize}
Compared to likelihood-based methods operating directly in image space, estimating the likelihood in a representation space with fewer dimensions results in more computationally efficient methods.
The evaluations demonstrate that likelihood for representations achieves results comparable to current state-of-the-art (SOTA) methods without requiring access to labeled data. Furthermore, by employing the labels of ID data and class conditional training (guidance) for the diffusion model, we are able to surpass the performance of most SOTA methods.

\section{Related Work} 

OOD detection was first introduced to recognize unseen data points unlikely to be part of the training distribution using statistical models \citep{bishop1994novelty}. Since then, it has gained significant research attention across multiple directions.

\paragraph{OOD detection with supervised classifiers} Many post-hoc methods derive distance functions from pre-trained classifiers, including MSP \citep{msp17iclr}, ODIN \citep{odin18iclr}, Mahalanobis distance \citep{mahananobis18nips}, Energy \citep{energyood20nips}, ReAct \citep{sun2021react}, ViM \citep{wang2022vim}, and Generalized Entropy \citep{liu2023gen}. These methods typically rely on supervised classifiers trained on ID data, using features from the penultimate layer, logits, or labels to define OOD scores. While widely adopted, their effectiveness relative to likelihood-based methods remains unclear.

\paragraph{DGM-based OOD detection} DGMs have been leveraged for OOD detection through reconstruction-based \citep{graham2023denoising}, likelihood-based \citep{nalisnick2019detecting, goodier2023likelihood}, and synthetic OOD data generation approaches \citep{du2022vos}. Unlike previous likelihood-based methods that model image pixels, we empirically study likelihood on image representation and benchmark with other lines of post-hoc methods. Compared to \citep{goodier2023likelihood}, our approach avoids the high computational cost associated with processing the high-dimensional image space.

\paragraph{OOD detection with self-supervised models} Self-supervised foundational models such as DINO~\citep{caron2021emerging} and DINOv2~\citep{oquab2023dinov2} have recently been explored for OOD detection \citep{ahmadian2024unsupervised, zhang2023openood}. However, most post-hoc methods \citep{msp17iclr, wang2022vim, liu2023gen} depend on labels or logits, making them unsuitable for self-supervised models. Only a few methods, such as Residual~\citep{wang2022vim} and KNN~\citep{sun2022out}, can be adapted for these models, highlighting a gap in OOD detection research as it was originally defined \citep{bishop1994novelty}.

\section{Representation likelihood estimation with diffusion models for OOD detection}\label{sec:method}

\subsection{\textbf{Preliminaries}}

\paragraph{OOD detection with ID labels} 
In every image classification problem, there is a predefined set of semantic categories that the model is expected to identify, which defines the ID images. We refer to this set of labels and the associated joint distribution as \(\idlabel\) and \(\id\), respectively, where \(\forall(\mathbf{x},y)\sim\id,\; y\in\idlabel\). In the open world, there are semantic groups that do not belong to the predefined and finite \(\idlabel\), forming the OOD space \(\oodlabel = \{y | y \notin \idlabel\}\) and \(\ood\). In this setup, OOD detection has two main goals \cite{zhang2023openood}. The first is to develop a discriminative model that accurately classifies ID samples drawn from \(\id\). The second goal is to develop a detector module (typically built upon the trained classifier) that accurately identifies whether an incoming image during the inference phase is ID or OOD.

\paragraph{OOD detection without ID labels} Access to annotated ID data points is not possible in many cases. As OOD detection is initially conceptualized in \citep{bishop1994novelty}, we can train a density model \(p_{\theta}(\mathbf{x})\) (where \(\theta\) represents the parameters) to approximate the true distribution of the training inputs \(p(\mathbf{x})\), given only \(\mathbf{x}\sim\iddata\) (where we use $\iddata$ for the marginal in-distribution of $\mathbf{x})$. Any $\mathbf{x}$ that has a sufficiently low density under \(p_{\theta}(\mathbf{x})\) is assigned to the \(\ood\) space if \(p_{\theta}(\mathbf{x})\) is a reasonably good estimation of \(p(\mathbf{x})\).

\subsection{Motivating observations}

\begin{figure}[t]
\centering
\includegraphics[width=\linewidth]{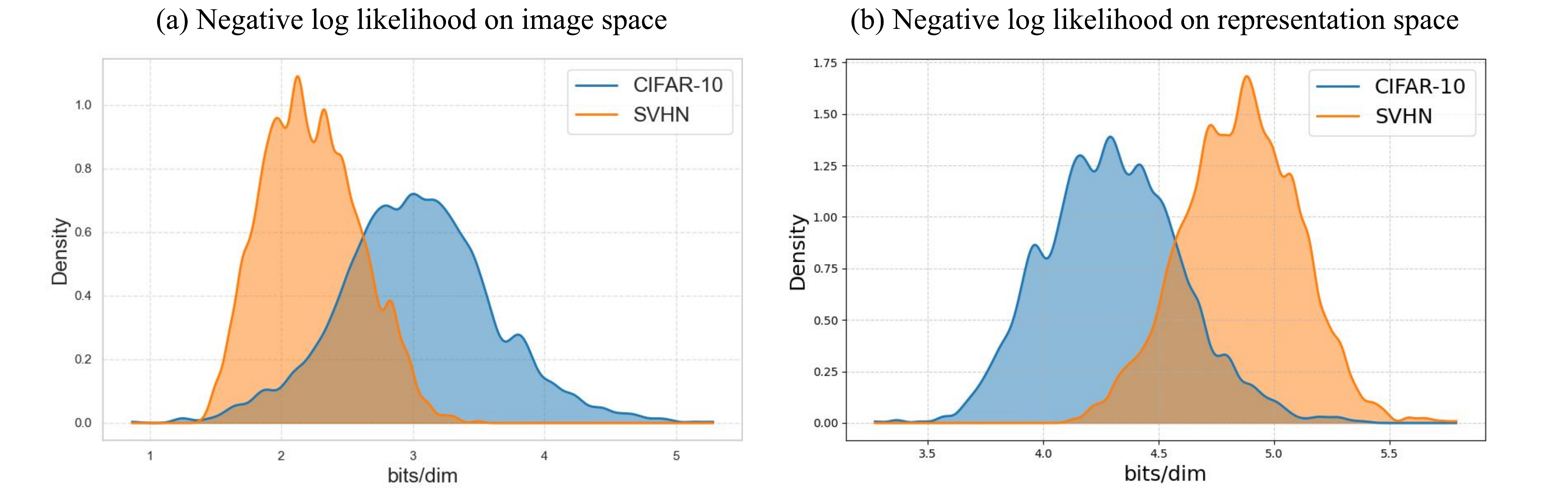}
\caption{Density of negative log likelihoods for CIFAR-10 (ID) vs SVHN (OOD) (a) diffusion model trained on CIFAR-10 images, (b) diffusion model trained on CIFAR-10 representations extracted from ResNet18.}
\label{fig:image_rep}
\end{figure}

Score-based diffusion models have gained attention for likelihood estimation due to their high expressiveness compared to other likelihood estimators \citep{song2020score}. However, similar to earlier flow-based models, as studied by \citep{nalisnick2018deep}, likelihood estimates from diffusion models remain ineffective to classify ID from OOD data in the image space. As illustrated in Fig.~\ref{fig:image_rep} (a), a diffusion model trained on the CIFAR-10 dataset continues to assign a lower negative log-likelihood value to the SVHN dataset, highlighting its limitations in OOD detection. In contrast, Fig.\ref{fig:image_rep}~(b) demonstrates that extracting representations from the penultimate layer of an encoder trained on CIFAR-10 using cross-entropy loss, such as ResNet18, enhances the effectiveness of likelihood-based OOD detection. However, the performance of likelihood estimation in the representation space compared to other state-of-the-art methods remains unclear. Thus, in this study we demonstrate that likelihood estimation is effective when applied to representations from both supervised and self-supervised encoders, as validated through comprehensive experiments on large-scale benchmarks. 

\subsection{\textbf{Pre-trained encoder}}
Given the image dataset \(\{\mathbf{x}_{i}\}_{i=0}^N\), representations \(\{\mathbf{z}_{i}\}_{i=0}^N\) are calculated by a pre-trained encoder \(\mathcal{E}\), \(\mathbf{z} = \mathcal{E}(\mathbf{x})\). We extract representations for both ID and OOD data using the same encoder, but only ID representations are used for training the diffusion model/likelihood estimator.

\subsection{OOD detection with representation likelihood estimation}

Models such as Generative Adversarial Networks (GANs) cannot calculate likelihood, while Denoising Diffusion Probabilistic Models (DDPMs) and Variational Autoencoders (VAEs) only provide a lower bound on the likelihood. However, using the (instantaneous) change of variables formula \cite{chen2018neural} with invertible generative models, such as normalizing flows, score-based diffusion models and flow matching, the likelihood can be computed exactly or estimated with high accuracy. 
In this work, we specifically use a score-based diffusion model \citep{song2020score} to estimate the likelihood, motivated by the flexibility and strong performance of this class of models for image generation. Since the model is trained on image representations, we refer to it as the Representation Diffusion Model (RDM) and its conditional version as ConRDM for brevity. 
Note that this is different from latent diffusion models \citep{rombach2022high} since we do not assume that we have access to a corresponding decoder. 
Training score-based diffusion models can be formulated as reverse-time stochastic differential equation (SDE) learning (for details, we refer to Appendix A), and the corresponding probability flow ODE \citep{maoutsa2020interacting} of such an SDE can be expressed as
\begin{equation}
    d \mathbf{z} = \{\mathbf{f}(\mathbf{z}, t) - \frac{1}{2} g(t)g(t)^T \nabla_{\mathbf{z}}\log p_t(\mathbf{z})\}dt, \quad t\in[0,1],
\end{equation}
where \(\mathbf{f}(\mathbf{z}, t)\) and \(g(t)\) are the drift and diffusion coefficients, respectively, from the underlying SDE and $p_t(\mathbf{z})$ is the marginal distribution of $\mathbf{z}$ at time $t$. Note that the ODE (in this formulation, following \citep{song2020score}) is initialized at time $1$ and runs backward in time, so $dt$ should be seen as a \textit{negative} infinitesimal time increment.
With the instantaneous change of variables formula \citep{chen2018neural}, denoting $f_{\theta}(\mathbf{z}, t) = \mathbf{f}(\mathbf{z}, t) - \frac{1}{2} g(t)g(t)^T s_\theta(\mathbf{z}, t)$, and assuming $\nabla_{\mathbf{z}}\log p_t(\mathbf{z}) \approx s_\theta(\mathbf{z}, t)$, we can compute the representation likelihood $p_0(\mathbf{z})$ using
\begin{equation}
\label{eqa:likelihood}
\log p_0(\mathbf{z}(0)) = \log p_1(\mathbf{z}(1)) + \int_0^1 \nabla \cdot f_{\theta}(\mathbf{z}(t), t)dt.
\end{equation}
In practice we follow \citep{song2020score} and make use of Hutchinson's trace estimator \citep{hutchinson1989stochastic} for estimating the divergence $\nabla \cdot f_{\theta}(\mathbf{z}(t), t)$.

We can numerically estimate the data likelihood without prior assumption of the data distribution. After training only on ID representations, Eqn.(\ref{eqa:likelihood}) is used to calculate likelihood for both ID and OOD data. We expect $\log p_0(\mathbf{z}_{ID}) > \log p_0(\mathbf{z}_{OOD})$
%
%
in general and fix a threshold \(\lambda\) such that we are able to make a decision for an input test image \(x\),
\begin{equation}
x \in
    \begin{cases}
      \mathcal{X}_{\text{ID}} & \text{if \(\log p_0(\mathbf{z}) \geq \lambda\)},\\
      \mathcal{X}_{\text{OOD}} & \text{if \(\log p_0(\mathbf{z}) < \lambda\)}.
    \end{cases}       
\end{equation}

\subsection{Leveraging labeled in-distribution data}

When ID labels are available, class-specific information can enhance OOD detection. Studies on diffusion models for density estimation and sampling indicate that conditioning the score function improves the samples quality while reducing their variance \cite{ho2022classifier}. Following this principle, a class-conditioned model is expected to assign higher likelihoods to high-confidence regions of ID classes in the representation space. To incorporate labels and estimate \( p(\mathbf{z} | c) \), where $c$ denotes the class conditioning, the model learns a class-conditioned diffusion score function \( s_\theta(\mathbf{z}, c, t) \). Inspired by classifier-free guidance \cite{ho2022classifier}, we introduce \( c \) as an additional input to the neural network. Here, we assume a pretrained encoder \( \mathcal{E} \) trained via supervised classification on ID data. Given an image \( \mathbf{x} \), we denote the classifier’s prediction as \( c = {\arg\max} \mathbf{Wz} + \mathbf{b}\), where \(\mathbf{W}, \mathbf{b}\) are the last MLP layer weights. Pseudo-code for training and detection is provided in Alg.~\ref{algo:training} and Alg.~\ref{algo:detection}, respectively.

\noindent
\begin{minipage}{0.5\linewidth}
\begin{algorithm}[H]
  \caption{Training}
  \label{algo:training}
  \scriptsize
  \SetAlgoNlRelativeSize{-1}
  \textbf{Input} Encoder \(\mathcal{E}\), training images \(\{\mathbf{x}^{id}_{i}\}_{i=0}^N\)  \\ 
  \textbf{Extract representation} \(\{\mathbf{z}^{id}_{i}\}_{i=0}^N\) by \(\mathbf{z} = \mathcal{E}(\mathbf{x})\)\\
  \textbf{If} class condition \textbf{then} \\
  ~~\(c = {\arg\max} \mathbf{Wz} + \mathbf{b}\) \\
  \textbf{Else} \(c = \emptyset\) \\
  \textbf{Train}  RDM \(\mathbf{D}(\mathbf{z}, c)\)\\
  
  \Return \(\mathbf{D}\)
\end{algorithm}
\end{minipage}
\begin{minipage}{0.5\linewidth}
\begin{algorithm}[H]
  \caption{Detection}
  \label{algo:detection}
  \scriptsize
  \SetAlgoNlRelativeSize{-1}

  \textbf{Input}  RDM \(\mathbf{D}\), encoder \(\mathcal{E}\), image \(\mathbf{x}\), threshhold \(\lambda\) \\ 
  \textbf{Extract representation} \(\mathbf{z}\) by \(\mathbf{z} = \mathcal{E}(\mathbf{x})\)\\
  \textbf{If} class condition \textbf{then} \\
  ~~\(c = {\arg\max} \mathbf{Wz} + \mathbf{b}\) \\
  \textbf{Else} \(c = \emptyset\) \\
  \textbf{Calculate} \(\log p(\mathbf{z}|c)\) \\

   \textbf{If} \(\log p(\mathbf{z}|c) \geq \lambda \),~~\(\mathbf{x} \in \mathcal{X}_{\text{ID}}\)\\\textbf{Otherwise} \(x \in  \mathcal{X}_{\text{OOD}}\) \\
   \vspace{1.3mm}
\end{algorithm}
\end{minipage}

\section{Experiments}
In this section, we present detailed experiment settings and evaluate likelihood for both \textit{OOD detection with ID labels} and \textit{OOD detection without ID labels} on a large-scale benchmark and a histopathology benchmark.

\subsection{Experiment settings}

\paragraph{Data and metrics} We use a large-scale OOD detection benchmark, utilizing ImageNet-1K as the ID dataset. As OOD data, we use four widely recognized datasets, including three far-OOD datasets: OpenImage-O, Texture and iNaturalist, and one near-OOD dataset: ImageNet-O, following the same evaluation protocol and dataset settings as in \citep{liu2023gen, wang2022vim}. For the histopathology benchmark, we use the PatchCamelyon (PCam) dataset \cite{Veeling2018-qh} and define non-tumor images as ID data and images containing tumor as OOD data, i.e. we perform unsupervised tumor detection. The histopathology benchmark represents a clinical application of OOD detection where labels of ID data are not available. We employ two conventional metrics to evaluate the OOD detection performance. The first is a threshold independent metric: Area Under the Receiver Operating Characteristic Curve (AUROC), where higher percentages reflect better performance. The second metric is the False Positive Rate at 95\% True Positive Rate (FPR95), with lower percentages indicating better performance.

\paragraph{Encoders} In our experiments, we use both supervised and self-supervised encoders. We use the same supervised encoders as in \citep{wang2022vim, liu2023gen} to allow for a fair comparison, as well as 5 self-supervised models, where 2 are trained on histopathology data. The details of the encoders are listed in Table \ref{tab:encoder}.

\begin{table}[t]
    \hspace{-3mm}
    \centering
    \resizebox{1.02\linewidth}{!}{
    \setlength{\tabcolsep}{0.5mm}
    \def\arraystretch{1.15}
    \begin{tabular}{lllccc}
        \toprule
        \textbf{Model}               & \textbf{Specification}       & \textbf{Architecture} & \textbf{Training Method} & \textbf{DIM} & \textbf{Dataset}\\
        \midrule
        BiT & BiT-S-R101x1                 & CNN                   & Cross Entropy                          & \(2048 \)          &ImageNet-21K \\

        RepVGG & RepVGG-b3                    & CNN                   & Cross Entropy                          & \(2560\)          &ImageNet-21K\\
        ResNet50d      & ResNet-50d                   & CNN                   & Cross Entropy                          & \(2048\)          &ImageNet-21K\\
        Swin      & Swin-B & Transformer           & Cross Entropy                 & \(1024\)          &ImageNet-21K\\
        ViT & ViT-B/16                     & Transformer           & Cross Entropy                 & \(768\)         &ImageNet-21K \\

        DeiT          & ViT-B/16        & Transformer           & Cross Entropy                          & \(768\)       &ImageNet-21K   \\
                MAE  & ViT-B/16                     &
        Transformer           & Self Supervised Learning   & 768  &ImageNet-1K\\
        DINO   & ViT-B/16                  &
        Transformer           & Self Supervised Learning   & 768  &ImageNet-1K\\
        DINOv2  & ViT-B/14                     &
        Transformer           & Self Supervised Learning   & 768  &LVD-142M\\

                Pathology-SSL  & ViT-S/16                     &
        Transformer           & Self Supervised Learning   & 384  &TCGA\&TULIP\\
        Uni  & ViT-L/16                     &
        Transformer           & Self Supervised Learning   & 1024 &Mass-100K \\
        
        \bottomrule
         
    \end{tabular}
    }

    \caption{
        Encoders used for extracting representations and their corresponding pre-training datasets and dimension (DIM)  on the representation vector $\mathbf{z}$.
    }
    \label{tab:encoder}
    \hspace{-100mm}
\end{table}

\paragraph{Details of the representation diffusion model} The RDM is parameterized by a time-dependent MLP with 12 residual blocks, following the architecture proposed by \citep{li2023self}. The model is trained using the AdamW optimizer, with a batch size of 4096 and a learning rate of 2e-3. Cosine annealing and gradient clipping are applied during training. Similar to \cite{song2020score}, we don't use likelihood weighting. For training, we use a sub-VP SDE \cite{song2020score} as the default, and no significant differences were observed when comparing Variance Preserving (VP) SDE to sub-VP SDE (details in Appendix A).

\paragraph{Likelihood estimation} The likelihood is calculated on the ImageNet-1K validation set (ID dataset) and across four different OOD datasets. Regarding the PCam dataset, the likelihood is computed only on the tumor and non-tumor test splits. For likelihood estimation, we solve Eq.~\ref{eqa:likelihood} with the RK45 ODE numerical integrators provided by \textit{torchdiffeq}\footnote{https://github.com/rtqichen/torchdiffeq}, where \textit{atol=1e-5} and \textit{rtol=1e-5}. For the divergence term, the Skilling-Hutchinson trace estimator \cite{hutchinson1989stochastic, skilling1989eigenvalues} is used. Unless otherwise mentioned, we use the same settings for all encoders. 

\paragraph{Computational efficiency and reproducibility} The training time of the diffusion model with 200 epochs is approximately 12 minutes on an RTX4090 GPU. Likelihood estimation has a throughput of 1500 image representations per second. The whole evaluation time on the ImageNet benchmark is approximately 25 seconds. We provide the source code of all training and evaluation implementations in the supplementary files.

\begin{table*}[t]
\centering
\resizebox{\textwidth}{!}{
\begin{tabular}{@{}lcccccccccc@{}}
\toprule
\textbf{Method} & \multicolumn{2}{c}{\textbf{OpenImage-O}} & \multicolumn{2}{c}{\textbf{Textures}} & \multicolumn{2}{c}{\textbf{iNaturalist}} & \multicolumn{2}{c}{\textbf{ImageNet-O}} & \multicolumn{2}{c}{\textbf{Average}} \\
 & {\footnotesize AUROC} $\uparrow$ & {\footnotesize FPR95} $\downarrow$ 
 & {\footnotesize AUROC} $\uparrow$ & {\footnotesize FPR95} $\downarrow$ 
 & {\footnotesize AUROC} $\uparrow$ & {\footnotesize FPR95} $\downarrow$ 
 & {\footnotesize AUROC} $\uparrow$ & {\footnotesize FPR95} $\downarrow$ 
 & {\footnotesize AUROC} $\uparrow$ & {\footnotesize FPR95} $\downarrow$ \\
\midrule
\multicolumn{11}{c}{\textbf{MAE}} \\
KNN\citep{sun2022out}                 & \textbf{60.54} & \textbf{89.03} & 89.04 & 41.51 & \textbf{48.02} & \textbf{97.69} & 68.64 & 81.20 & \textbf{66.56} & 77.36 \\
Residual w/o offset\citep{wang2022vim} & 59.52 & 89.22 & \textbf{90.33} & \textbf{38.90} & 42.47 & 98.87 & \textbf{69.60} & \textbf{79.85} & 65.48 & \textbf{76.71} \\
\textbf{RDM}                           & 58.15 & 91.50 & 89.06 & 43.80 & 41.44 & 99.20 & 66.40 & 86.25 & 63.76 & 80.19 \\
\multicolumn{11}{c}{\textbf{DINO}} \\
KNN\citep{sun2022out}                 & 85.26 & 65.25 & 94.15 & 25.39 & 88.30 & 67.62 & 81.55 & 74.70 & 87.31 & 58.23 \\
Residual w/o offset\citep{wang2022vim} & \textbf{87.57} & \textbf{54.77} & \textbf{97.84} & \textbf{11.10} & \textbf{92.71} & \textbf{42.76} & \textbf{81.98} & \textbf{68.40} & \textbf{90.02} & \textbf{44.25} \\
\textbf{RDM}                           & 85.68 & 64.73 & 96.59 & 17.17 & 86.67 & 70.98 & 79.80 & 73.90 & 87.18 & 56.69 \\
\multicolumn{11}{c}{\textbf{DINOv2}} \\
KNN\citep{sun2022out}                 & \textbf{95.05} & \textbf{25.66} & 91.65 & 35.33 & 99.06 & 3.47  & \textbf{86.67} & \textbf{57.55} & 93.10 & \textbf{30.50} \\
Residual w/o offset\citep{wang2022vim} & 92.61 & 35.53 & \textbf{93.60} & 33.41 & \textbf{99.32} & \textbf{1.74} & 83.23 & 70.40 & 92.19 & 35.26 \\
\textbf{RDM}                           & 94.06 & 31.07 & 93.32 & \textbf{32.50} & 99.30 & 1.83 & 85.97 & 63.30 & \textbf{93.16} & 32.17 \\
\bottomrule
\end{tabular}
}
\caption{\textbf{OOD detection with self-supervised encoder:} AUROC and FPR95 are reported as percentages. Results for MAE, DINO and DINOv2 with ImageNet-1K as ID data and four OOD datasets: OpenImage-O, Textures, iNaturalist, and ImageNet-O. Since logits are not available, we only compare with KNN \cite{sun2022out} and Residual \cite{wang2022vim}. The best method is marked in bold.}
\label{table:ssl}
\end{table*}

\subsection{Evaluation with self-supervised encoders}

The representation likelihood provides flexibility to detect OOD data with self-supervised encoders when label information of ID data is not available. We evaluate our approach on both the large-scale benchmark and the histopathology task with various self-supervised encoders and compare its performance against other methods that do not require labels.

\paragraph{ImageNet Benchmark} As illustrated in Table \ref{table:ssl}, RDM is compared to two other label-free methods, Residual \cite{sun2022out} and KNN \cite{wang2022vim}, across three different self-supervised encoders. Likelihood on image representations shows competitive performance, particularly when paired with the best performing encoder, DINOv2. Overall, the results are mixed across the three methods; however, the differences in performance between the encoders are more significant than the differences between the OOD detection methods. We find the optimal \(k = 50\) for KNN, which is selected from \(k = \{1, 10, 20, 50, 100, 200, 500, 1000, 3000, 5000\}\) \citep{sun2022out}. However, determining the optimal \(k\) requires a calibration dataset, and the optimal value may vary across different datasets or representations. The Residual score is used as part of ViM \cite{wang2022vim}. In the original implementation, an offset \(\mathbf{o} = (\mathbf{W}^T)^+\mathbf{b}\) is subtracted to ensure the results are unbiased. However the offset is from the penultimate layer of a supervised encoder and we simply set it to 0.

\begin{table}[h]
\centering
\resizebox{0.8\linewidth}{!}{%
\setlength{\tabcolsep}{0.5mm}
\def\arraystretch{1.1}
\begin{tabular}{@{}lcccccc@{}}
\toprule
\textbf{Method} & \multicolumn{2}{c}{\textbf{Uni~\cite{chen2024towards}}} & \multicolumn{2}{c}{\textbf{Pathology-SSL~\citep{kang2023benchmarking}}} & \multicolumn{2}{c}{\textbf{DINO~\cite{caron2021emerging}}} \\
 & {\footnotesize AUROC}\,$\uparrow$ & {\footnotesize FPR95}\,$\downarrow$ & {\footnotesize AUROC}\,$\uparrow$ & {\footnotesize FPR95}\,$\downarrow$ & {\footnotesize AUROC}\,$\uparrow$ & {\footnotesize FPR95}\,$\downarrow$ \\
\midrule
KNN  \citep{sun2022out}          & 88.36   & 50.74   & \textbf{85.28}   & \textbf{55.49}   & 60.77   & 98.17     \\
Residual w/o offset \cite{wang2022vim}   & 89.05   & 48.09   & \(75.78^*\)   & \(86.59^*\)   & 61.14   & 97.72       \\
\textbf{RDM}                & \textbf{89.38} & \textbf{45.25} & 81.49 & 68.54 & \textbf{61.29} & \textbf{96.71} \\
\bottomrule
\end{tabular}
}
\caption{\textbf{OOD detection with histopathology data:} AUROC and FPR95 are reported as percentages. \(^*\) The official Residual implementation is not defined for representation dimensions below 512, so the same ratio is used as in the DINO representation. The best method is marked in bold.}
\label{table:histo}
\end{table}
\paragraph{PCam benchmark} We also present experiments on the PCam dataset in Table~\ref{table:histo} using three self-supervised encoders: two pre-trained on histopathology data and one pre-trained on natural image data. The results show similarly mixed performance as seen in Table \ref{table:ssl}, with the differences between encoders being larger than the differences between methods.  Residual uses the smallest half to one-third of the principal subspace in the official implementation. However, the optimal number of principal components for Pathology-SSL representations spanned nearly the entire space (details in Appendix B). The diffusion model likelihood estimator, by contrast, is not sensitive to hyperparameters, as we use the same settings across all training on different representations.

\subsection{Evaluation with supervised encoders}
We now shift our attention to supervised encoders.
We present results using ViT (which achieved the highest average OOD detection performance among all encoders) in Table~\ref{tab:vit}, with results for other encoders provided in Table~\ref{tab:other}.
We compare the likelihood-based RDM and ConRDM with a comprehensive collection of methods from the literature.


\begin{table*}[h!]
    \centering
    \resizebox{\textwidth}{!}{
\begin{tabular}{@{}llccccccccccc@{}}
\toprule
\textbf{Method} &\textbf{Source} & \multicolumn{2}{c}{\textbf{OpenImage-O}}& \multicolumn{2}{c}{\textbf{Textures}} & \multicolumn{2}{c}{\textbf{iNaturalist}} & \multicolumn{2}{c}{\textbf{ImageNet-O}} & \multicolumn{2}{c}{\textbf{Average}}\\
 & & {\footnotesize AUROC} $\uparrow$ & {\footnotesize FPR95} $\downarrow$ & {\footnotesize AUROC} $\uparrow$ & {\footnotesize FPR95} $\downarrow$ & {\footnotesize AUROC} $\uparrow$ & {\footnotesize FPR95} $\downarrow$  & {\footnotesize AUROC} $\uparrow$ & {\footnotesize FPR95} $\downarrow$  & {\footnotesize AUROC} $\uparrow$ & {\footnotesize FPR95} $\downarrow$ \\
\midrule

\multicolumn{12}{c}{\textbf{With Label Information}}\\
{MSP}~\citep{msp17iclr}                     & prob         & 92.53   & 34.18   & 87.10   & 48.55   & 96.11   & 19.04   & 81.86   & 64.85   & 89.40   & 41.65   \\
{Energy}~\citep{energyood20nips}            & logit        & 97.11   & 14.04   & 93.39   & 28.22   & 98.66   & 6.16    & 90.46   & 41.30   & 94.90   & 22.43   \\
{ODIN}~\citep{odin18iclr}                   & prob+grad    & 96.86   & 15.68   & 93.01   & 30.60   & 98.57   & 6.58    & 89.85   & 44.15   & 94.57   & 24.25   \\
{MaxLogit}~\citep{hendrycks2019scaling}     & logit        & 96.87   & 15.68   & 93.01   & 30.60   & 98.57   & 6.58    & 89.85   & 44.15   & 94.57   & 24.25   \\
{KL Matching}~\citep{hendrycks2019scaling}  & prob         & 93.80   & 28.49   & 88.76   & 44.09   & 96.88   & 14.79   & 84.12   & 55.70   & 90.89   & 35.77   \\
{GEN}~\citep{liu2023gen}                   &  logit   & 96.60   & 17.13   & 92.35   & 34.01   & 98.63   & 5.83    & 89.67   & 47.60   & 94.31   & 23.14   \\
{ReAct}~\citep{sun2021react}                 & feat+logit   & 97.38   & 13.50   & 93.34   & 28.49   & 99.00   & 4.31    & 90.71   & 42.60   & 95.11   & 22.22   \\
{Mahalanobis}~\citep{mahananobis18nips}     & feat+label   & 97.48   & 13.54   & 94.24   & 25.17   & 99.54   & 2.12    & 92.81   & 36.95   & 96.02   & 19.45   \\
{ViM}~\citep{wang2022vim}                   & feat+logit   & \textbf{97.61}   & \textbf{12.61}   & 95.34   & 20.31   & 99.41   & 2.60    & 92.55   & 36.75   & 96.23   & 18.07   \\

\textbf{ConRDM}                         & feat+logit   & 97.37   & 14.16   & \textbf{95.41}   & \textbf{19.07}   & \textbf{99.49}   & \textbf{2.21}    & \textbf{93.15}   & \textbf{32.80}   & \textbf{96.35}   & \textbf{17.06}   \\
\midrule
\multicolumn{12}{c}{\textbf{Without Label Information}}\\
{Residual}~\citep{wang2022vim}              & feat         & 92.72   & 32.63   & 92.21   & 33.80   & 98.57   & 6.63    & 88.23   & 47.85   & 92.93   & 30.23   \\
{Residual w/o offset}~\citep{wang2022vim}
& feat         & 91.87   & 36.38   & 92.20   & 33.84   & 98.57   & 6.64    & 88.23   & 47.90   & 92.71   & 31.19   \\
{KNN}~\citep{sun2022out}                    & feat         & 93.54   & 38.92   & 92.95   & 29.40   & 94.68   & 35.65   & 88.86   & 52.80   & 92.51   & 39.20   \\
\textbf{RDM}                            & feat         & \textbf{95.98}   & \textbf{21.59}   & \textbf{94.25}   & \textbf{25.23}   & \textbf{99.12}   & \textbf{4.35}    & \textbf{91.41}   & \textbf{40.15}   & \textbf{95.18}   & \textbf{22.83}   \\
\bottomrule
\end{tabular}
}
\caption{\textbf{OOD detection with supervised encoder:} AUROC and FPR95 are reported as percentages. The ID dataset is ImageNet-1K, while the OOD datasets are OpenImage-O, Texture, iNaturalist, and ImageNet-O. The supervised ViT-B/16 encoder model is used for representation extraction. The best method is marked in bold. Source refers to the information that each method requires from the encoder network: feat (feature representations), prob (softmax probabilities), or logits.} 
\label{tab:vit}
\end{table*}

\begin{table}[h]
    \hspace{-3mm}
    \centering
    \resizebox{1.026\linewidth}{!}{
    \setlength{\tabcolsep}{0.5mm}
    \def\arraystretch{1.1}
\begin{tabular}{@{}lcccccccccc@{}}
\toprule
\textbf{Method}  & \multicolumn{2}{c}{\textbf{Swin}} & \multicolumn{2}{c}{\textbf{DeiT}} & \multicolumn{2}{c}{\textbf{RepVGG}} & \multicolumn{2}{c}{\textbf{ResNet50d}} & \multicolumn{2}{c}{\textbf{BiT}}\\
 & {\footnotesize AUROC}\,$\uparrow$ & {\footnotesize FPR95}\,$\downarrow$ & {\footnotesize AUROC}\,$\uparrow$ & {\footnotesize FPR95}\,$\downarrow$ & {\footnotesize AUROC}\,$\uparrow$ & {\footnotesize FPR95}\,$\downarrow$  & {\footnotesize AUROC}\,$\uparrow$ & {\footnotesize FPR95}\,$\downarrow$  & {\footnotesize AUROC}\,$\uparrow$ & {\footnotesize FPR95}\,$\downarrow$ \\
\midrule

\multicolumn{10}{c}{\textbf{With Label Information}}\\
{MSP}~\citep{msp17iclr}                             & 87.57    & 43.44    & 79.48    & 66.43    & 78.10    & 70.55    & 77.99    & 67.96    & 77.25 & 77.83 \\
{Energy}~\citep{energyood20nips}                  & 87.77    & 35.08    & 72.80    & 70.14    & 76.38    & 78.99    & 71.08    & 78.39    &78.48 &79.68\\
{ODIN}~\citep{odin18iclr}                     & 88.00    & 36.58    & 77.13    & 63.92    & 77.72    & 72.68    & 75.27    & 68.56    &79.24 &78.63\\
{MaxLogit}~\citep{hendrycks2019scaling}            & 88.40    & 35.28    & 76.79    & 64.49    & 77.56    & 73.50    & 75.39    & 69.34    &79.27 &78.46\\
{KL Matching}~\citep{hendrycks2019scaling}          & 88.87    & 46.99    & 83.49    & 64.80    & 81.35    & 61.65    & 82.72    & 64.41    &83.63 &55.62\\
{GEN}~\citep{liu2023gen}          & 91.46    & 32.28    & 84.61    & \textbf{59.68}    & 81.33    & 66.00    & 82.75    & 62.08    &80.00 & 81.00\\
{ReAct}~\citep{sun2021react}                  & 90.17    & 31.36    & 77.37    & 67.00    & 49.14    & 98.96    & 82.93    & 58.63    &84.53 &61.38\\
{Mahalanobis}~\citep{mahananobis18nips}        & 92.16    & 40.39    & 85.03    & 73.18    & 86.07    & 59.39    & 88.33    & 55.70    &86.62 &53.34\\
{ViM}~\citep{wang2022vim}                     & \textbf{94.11}    & \textbf{31.04}    & \textbf{85.25}    & 69.95    & \textbf{87.81}    & \textbf{50.50}    & \textbf{89.22}    & \textbf{52.61}    & \textbf{90.91} & \textbf{41.46}\\
\textbf{ConRDM}                           & 91.97    & 42.12    & 83.79    & 77.94    & 83.35    & 64.58    & 84.12    & 66.20    & 82.83    & 57.68 \\
\midrule
\multicolumn{10}{c}{\textbf{Without Label Information}}\\
{Residual}~\citep{wang2022vim}                      & \textbf{92.88}    & \textbf{37.38}    & 84.15    & 74.13    & 84.19    & 59.00    & 87.01    & 58.55   &84.14 &56.23 \\
{Residual w/o offset}  & 92.81 & 37.77 & 84.16 & 74.28 & 83.98 & 59.42 & 86.72 & 59.27    &84.05 & 56.26                   \\
{KNN}~\citep{sun2022out} &92.16 & 38.72 & \textbf{85.53} & \textbf{72.16} & \textbf{87.80} & \textbf{51.80} & \textbf{89.70} & \textbf{48.34} & \textbf{85.16} & \textbf{53.93}                         \\
\textbf{RDM}                                   &91.58 & 43.38 &82.93 & 79.44 & 82.01 & 65.26 &83.22 & 66.78  &81.72 & 58.56 \\
\bottomrule
\end{tabular}
}
\caption{\textbf{OOD detection with other supervised encoders:} AUROC and FPR95 are reported as percentages for 5 more supervised classifiers. The results are averaged on OpenImage-O, Texture, iNaturalist, and ImageNet-O for each classifier.} 
    \label{tab:other}
\end{table}

\paragraph{OOD detection with ViT} In general, both RDM and ConRDM demonstrate superior performance across the benchmarks. In the label-free setup, RDM consistently outperforms the second-best method, Residual, by 4.11\% in AUROC and 14.79\% in FPR95. Specifically, RDM achieves notable improvements in challenging OOD datasets like ImageNet-O, where it reduces FPR95 to 40.15\% compared to Residual's 47.90\%. On the Textures dataset, RDM maintains competitive performance with an FPR95 of 25.23\%, improving over Residual’s 33.84\%. With additional label information, ConRDM surpasses all methods, achieving the highest average AUROC of 96.35\% and FPR95 of 17.06\%. Overall, the results indicate that RDM offers robust OOD detection capabilities, with consistent performance gains across diverse datasets. Conditional likelihood further improves OOD detection when label information is available, making it a promising method for OOD detection in both label-free and label-augmented settings.

\begin{figure}[h]
\centering
\includegraphics[width=0.95\linewidth,trim={3.4cm 3mm 3.75cm 1.5cm},clip]{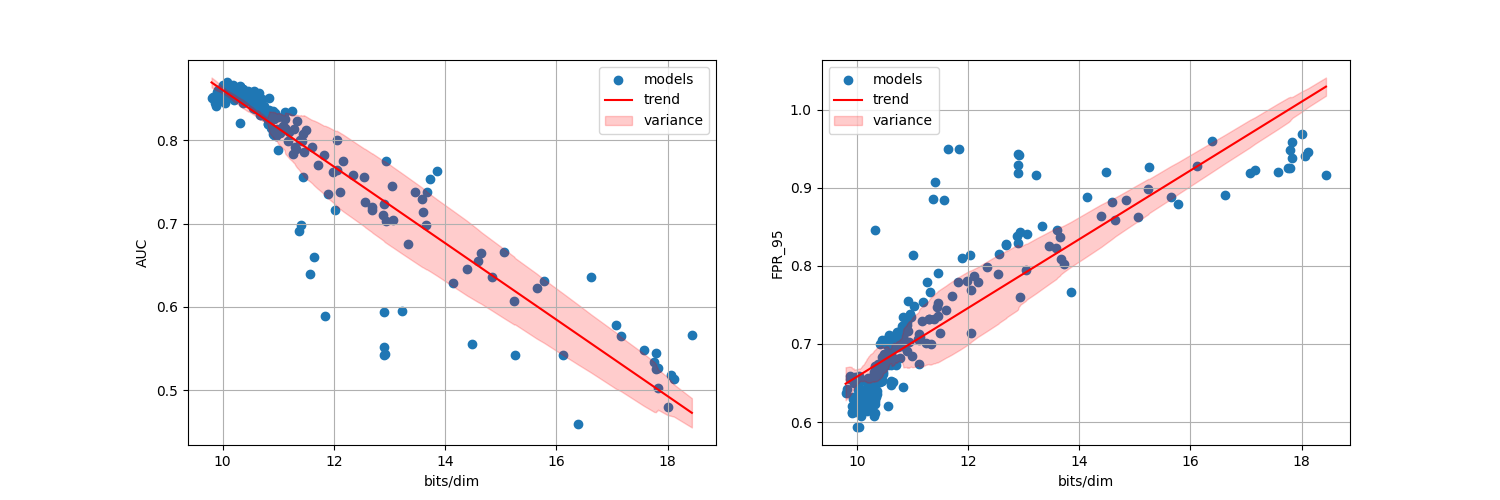}
\caption{More precise likelihood estimation (the lower bits/dim the better) of ID representations leads to more accurate OOD detection. The representations used in this analysis are extracted from DINOv2.}
\label{fig:aucvslike}
\end{figure}

\paragraph{OOD Detection with Alternative Encoders}
Table \ref{tab:other} presents a comprehensive comparison of OOD detection performance across various encoders, with results averaged over all OOD datasets. Representations extracted from CNNs, including BiT, RepVGG, and ResNet50d, exhibit asymmetrical representation spaces due to ReLU and are characterized by higher dimensionality compared to transformer-based encoders, with feature dimensions of 2048, 2560, and 2048, respectively. Likelihood-based methods demonstrate performance on par with state-of-the-art OOD detection approaches. ConRDM achieves a 1\% improvement over RDM by leveraging label information. However, its performance is slightly lower on CNN-based encoders, such as RepVGG, ResNet50d, and BiT, potentially due to the increased dimensionality, which may necessitate larger network expressiveness for effective likelihood estimation.

\subsection{Representation space analysis}
\label{sec:repanalysis}

In this part, we analyze the relationship between the accuracy of the likelihood estimation in the representation space and the OOD detection performance. Zhang et al. \citep{zhang2021understanding} found that sometimes DGMs with better likelihood estimation can perform worse in terms of OOD detection in certain distributions. We empirically demonstrate that this is not the case in the representation space, see Fig.~\ref{fig:aucvslike}. We find that more accurate likelihood estimation (low average bits/dim in ID validation dataset) leads to better OOD detection. Each point in Fig.~\ref{fig:aucvslike} indicates a model trained with a randomly specified setting, selected from different selections of the number of network layers, learning rate, and training epoch.

\paragraph{Discussion} Likelihood-based methods have traditionally been considered ineffective for OOD detection, particularly in raw image space, where models often assign anomalously high likelihoods to OOD samples. However, our study demonstrates that when applied to well-structured representations from high-quality encoders, likelihood estimation becomes a powerful and reliable approach for OOD detection. Through extensive experiments, we show that both supervised and self-supervised encoders provide effective feature spaces where likelihood can differentiate ID from OOD samples. In particular, likelihood is highly practical for scenarios where labeled data is unavailable.
 

\section{Conclusion}

We revisited likelihood-based OOD detection using a score-based diffusion model in the representation spaces provided by pre-trained encoders. Our results show that likelihood in such spaces performs comparably to SOTA methods without requiring labels for the ID images, making it a strong contender for OOD detection with, e.g., self-supervised encoders. When a supervised encoder is available, the model can be trained with class information as conditional likelihood. This formulation outperforms most SOTAs with the ViT encoder, achieving an average AUROC of 96.35 on the large-scale OOD detection benchmark with ImageNet-1K as ID data.
Our results show that likelihood-based methods can indeed be successfully used for OOD detection. While these methods often face criticism for their poor performance in image space, our approach demonstrates its effectiveness when applied in representation space. 
From these results, we argue that likelihood-based OOD detection, which to a large extent has been replaced by post-hoc methods, remains a powerful strategy when applied to representations from foundational models.

One of the advantages of likelihood-based OOD detection is the absence of detection related hyper-parameters, while one limitation is the sensitivity to the representation used, e.g., the RDM method does not perform well on representations from convolutional encoders.
For future work, we envision that further improvements can be made both by improving the likelihood estimator and the representations. Although the score-based diffusion model excels at density estimation, it would be interesting to test other generative models that allow precise likelihood estimation. As for the representations, a promising future research direction is to fine-tune the encoder to promote characteristics that benefit OOD detection performance.

\section*{Acknowledgement}

This work was partially supported by the Wallenberg AI, Autonomous Systems and
Software Program (WASP) funded by the Knut and Alice Wallenberg Foundation, and the Zenith career development program at Linköping University.

%
%
%
{\small
\bibliographystyle{splncs04}
\bibliography{reference}
}

\renewcommand\thesection{\Alph{subsection}}

\section*{Appendix}
\subsection{Score-based representation diffusion models}

\begin{figure}[H]
\centering
\includegraphics[width=0.9\textwidth]{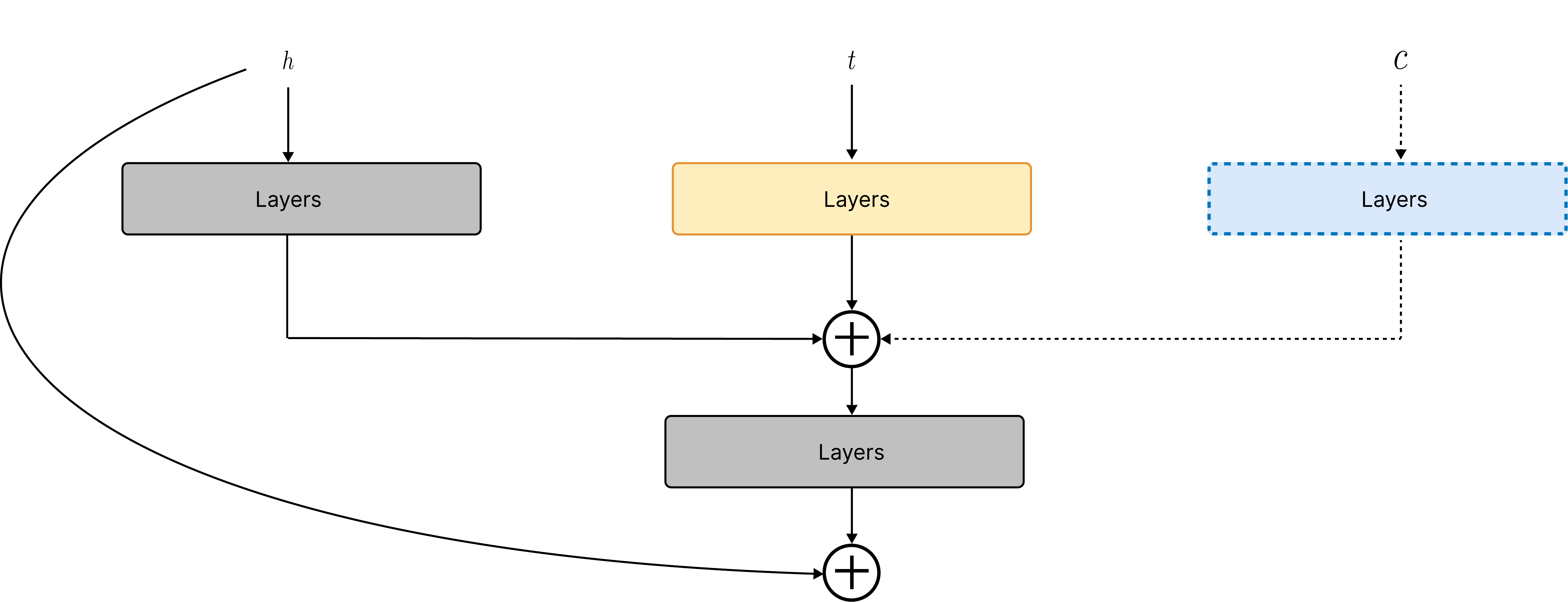}
\caption{Structure of each residual block. \(h, t, c\) are hidden feature, time and class condition, respectively.}
\label{fig:block}
\end{figure}

The model used in our study is a score-based diffusion model, employed to estimate the density of the extracted representation \(\mathbf{z}\). The forward-time diffusion process projects the representation distribution \(p_0(z)\) to the noise distribution \(p_1(z)\), which follows a stochastic differential equation (SDE), 
\begin{equation}
\label{eqn:sde}
d \mathbf{z} = \mathbf{f}(\mathbf{z}, t) d t + g(t) d \mathbf{w},
\end{equation}
where $\mathbf{w}$ is the Brownian motion, $\mathbf{f}(\mathbf{z}, t)$ is a vector-valued function, and $g(t)$ is a scalar function known as the diffusion coefficient. Sampling from the prior noise distribution $p_t$ and reversing the diffusion process, we can obtain a sample from the data distribution $p_0$. This reverse diffusion process is given by the reverse-time SDE,
\begin{equation}
  d\mathbf{z} = [\mathbf{f}(\mathbf{z}, t) - g^2(t)\nabla_{\mathbf{z}}\log p_t(\mathbf{z})] dt + g(t) d\bar{\mathbf{w}}.
\end{equation}
A time-dependent score network $s_\theta(\mathbf{z}, t)$ can be trained to approximate the score $\nabla_\mathbf{z} \log p_t(\mathbf{z})$, using the weighted sum of denoising score matching objectives,
\begin{equation}
\min_\theta \mathbb{E}_{t\sim \mathcal{U}(0, 1)} [\lambda(t) \mathbb{E}_{\mathbf{z}(0) }\mathbf{E}_{\mathbf{z}(t)}[ \|s_\theta(\mathbf{z}(t), t)\\- \nabla_{\mathbf{z}(t)}\log p_{0t}(\mathbf{z}(t) \mid \mathbf{z}(0))\|_2^2]].
\end{equation}
Where $\mathbf{z}(0) \sim p_0(\mathbf{z})$, $\mathbf{z}(t) \sim p_{0t}(\mathbf{z}(t) \mid \mathbf{z}(0))$, $\mathcal{U}(0,1)$ is a uniform distribution over $[0, 1]$, $p_{0t}(\mathbf{z}(t) \mid \mathbf{z}(0))$ denotes the transition probability from $\mathbf{z}(0)$ to $\mathbf{z}(t)$, and $\lambda(t)$ denotes a positive weighting function.

The $s_\theta(\mathbf{z}, t)$ is parameterized by a residual MLP network, each block is shown in Fig. \ref{fig:block}. The class condition layer is only used for ConRDM and encoded by Fourier features up to 256 dimension.
In practice, the training objective of our model is determined by selecting one of the Variance Exploding (VE), Variance Preserving (VP), or subVP forms for the SDE, for \(\sigma_{min}, \sigma_{max}\) and \(\beta_{min}, \beta_{max}\), we use the default setting, as outlined in Table~\ref{tab:SDE}.

\begin{table}[H]
    \centering
    \resizebox{0.9\textwidth}{!}{
    \begin{tabular}{lcc}
        \toprule
        \textbf{SDEs}               & \textbf{Formulation}        & \textbf{Setting}   \\
        \midrule
        VE SDE & \(
    d\mathbf{z} = \sqrt{\frac{\mathrm{d} \left[ \sigma^2(t) \right]}{\mathrm{d} t}} d\mathbf{w}
\)                                 & \(\sigma_{min}=0.01, \sigma_{max}=50\)                           \\
        VP SDE & \(d\mathbf{z} = -\frac{1}{2}\beta(t) \mathbf{z}~ dt + \sqrt{\beta(t)} ~d \mathbf{w}\)                              & \(\beta_{min}=0.2, \beta_{max}=20\)                 \\
        subVP SDE  & \(d\mathbf{z} = -\frac{1}{2}\beta(t) \mathbf{z}~ d t + \sqrt{\beta(t)(1 - e^{-2\int_0^t \beta(s)d s})} d \mathbf{w}\)  & \(\beta_{min}=0.2, \beta_{max}=20\)\\
        
        \bottomrule
    \end{tabular}
    }
    \caption{
        Formulation of 3 different SDEs, including hyperparameters that are used in our method. 
    }
    \label{tab:SDE}
\end{table}

\subsection{2D dataset experiments}

\begin{figure}[H]
\centering
\includegraphics[width=0.8\textwidth]{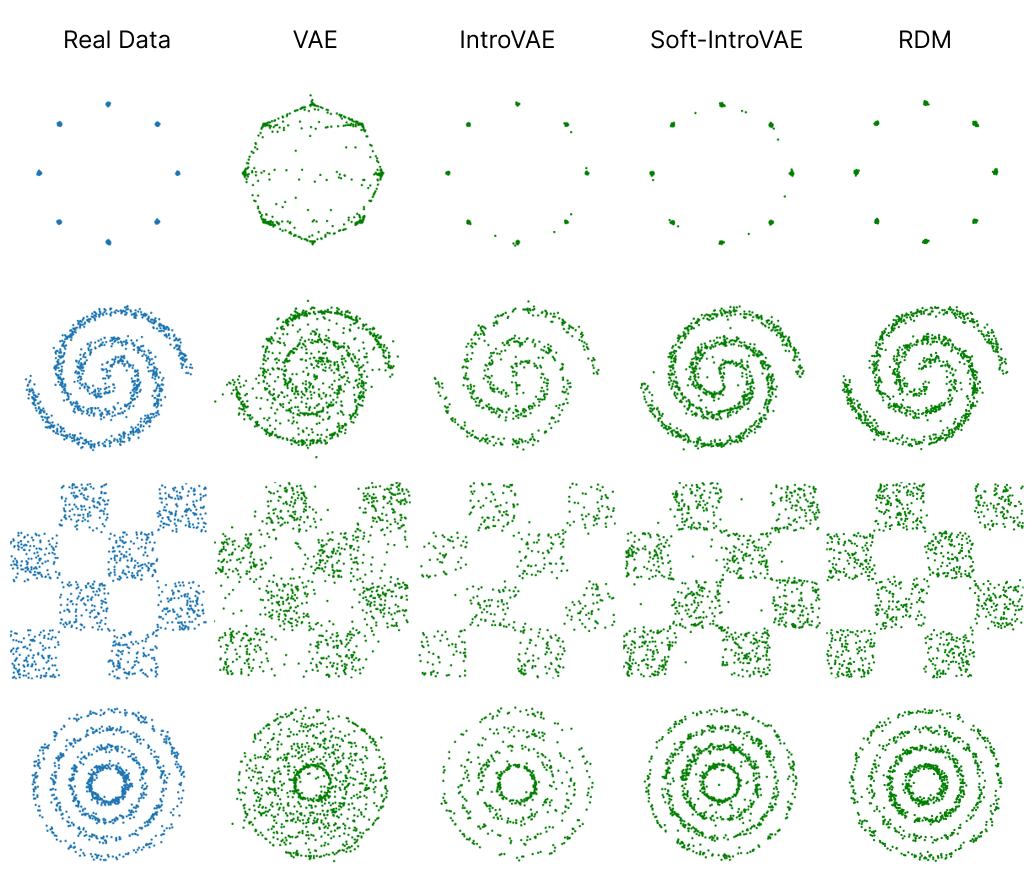}
\caption{Samples for 4 different 2D datasets. Results for VAEs are from \cite{daniel2021soft}}.
\label{fig:2d}
\end{figure}

We evaluate the quality of data distribution estimation on four 2D datasets: 8 Gaussians, Spiral, Checkerboard, and Rings, as shown in Table \ref{tab:toy_exp}. The results for VAEs are taken from \cite{daniel2021soft}, and following their methodology, we conduct our experiments using five different random seeds. To ensure a fair comparison, the complexity of our network architecture is kept on a similar scale to that used in \cite{daniel2021soft}, and we train for the same 30,000 iterations and sample 5000 points. We did not perform hyperparameter optimization, and use the same settings as in the high-dimensional representation experiments. The 2D points are sampled using an ODE sampler \cite{song2020score} as illustrated in Fig. \ref{fig:2d}. RDM performs particularly well in high-frequency data sampling, such as the 8 Gaussians dataset, where it achieves significantly lower KL-divergence compared to other methods.

\begin{table}[H]
\begin{center}
\begin{scriptsize}
\resizebox{0.8\textwidth}{!}{
\begin{tabular}{lccccccc}
    \toprule
    & & \textbf{VAE \citep{kingma2013auto}} & \textbf{IntroVAE} \citep{huang2018introvae} & \textbf{Soft-IntroVAE} \citep{daniel2021soft} & \textbf{RDM}\\
    \hline
    \multirow{3}{*}{8 Gaussians}
    &KL & 6.72$\pm$0.46 & 2.53$\pm$1.07 & 1.25$\pm$0.11 & \textbf{0.39$\pm$0.11}\\
    &JSD & 16.04$\pm$0.3 & 1.67$\pm$0.46 & 0.96$\pm$0.15 & \textbf{0.61$\pm$0.10}\\
    \hline
    \multirow{3}{*}{Spiral}
    &KL & 9.8$\pm$0.48 & 8.38$\pm$0.45 & 8.13$\pm$0.3 & \textbf{7.06$\pm$0.36}\\
    &JSD & 4.89$\pm$0.05 & 3.58$\pm$0.04 & 3.37$\pm$0.04 & \textbf{3.08$\pm$0.03}\\
    \hline
    \multirow{3}{*}{Checkerboard}
    &KL & 20.91$\pm$0.45 & 19.03$\pm$0.34 & 20.27$\pm$0.21 & \textbf{17.36$\pm$0.25}\\
    &JSD & 9.78$\pm$0.04 & 9.07$\pm$0.1 & 9.06$\pm$0.15 & \textbf{8.37$\pm$0.14}\\
    \hline
    \multirow{3}{*}{Rings}
    &KL & 13.16$\pm$0.55 & 10.21$\pm$0.49 & 9.18$\pm$0.33 & \textbf{7.74$\pm$0.39}\\
    &JSD & 7.26$\pm$0.07 & 4.24$\pm$0.11 & 4.13$\pm$0.09 & \textbf{3.60$\pm$0.11}\\
    \hline
\end{tabular}
}
\end{scriptsize}
\end{center}
\caption{Experiments on 4 types of 2D toy datasets. KL-divergence and
Jensen–Shannon-divergence (JSD) are evaluated. Results for VAEs are from \cite{daniel2021soft}.}
\label{tab:toy_exp}
\end{table}

\subsection{Residual setting on PCam dataset}

\begin{figure}[H]
\centering
\includegraphics[width=0.8\textwidth]{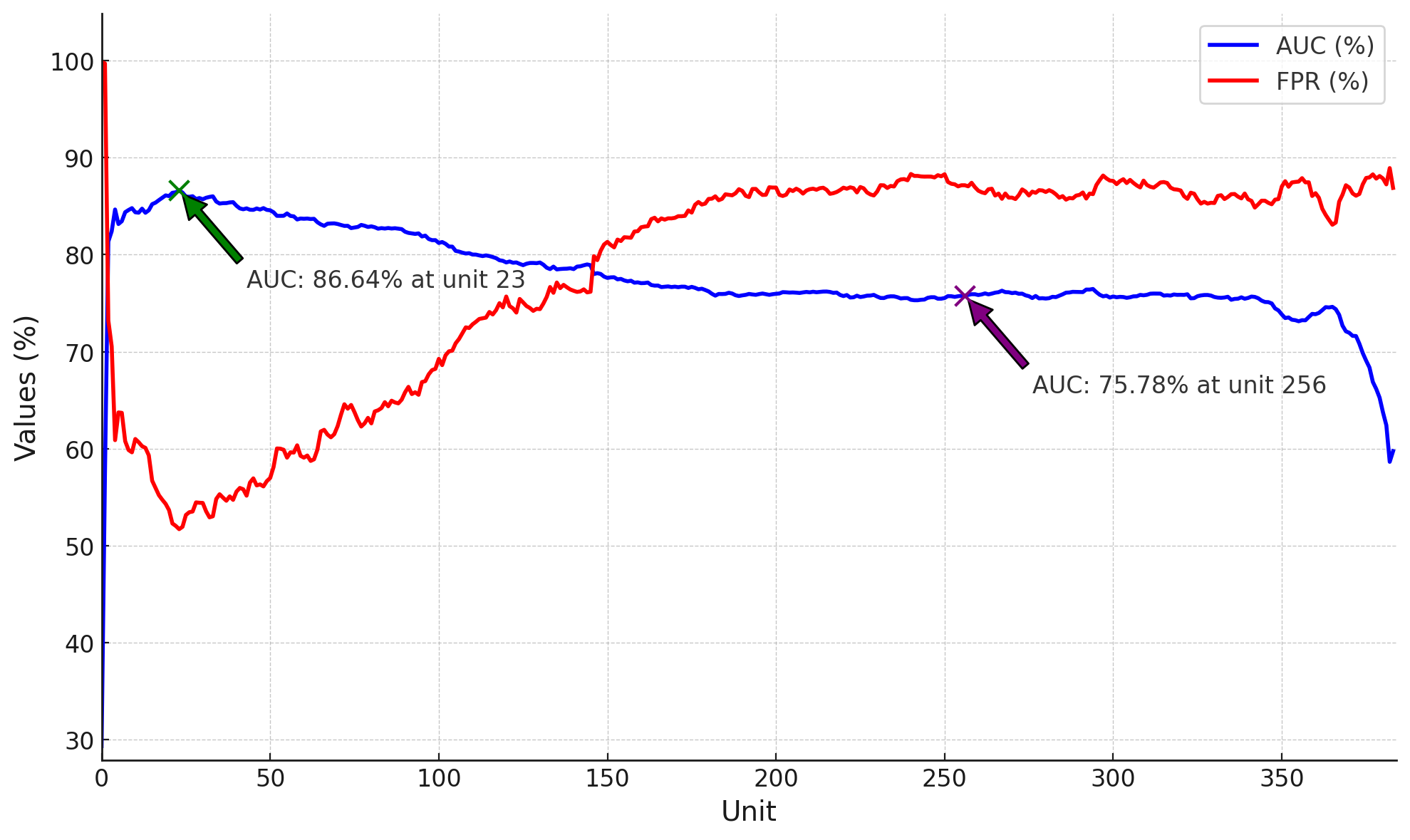}
\caption{For the PCam dataset, AUC and FPR95 are reported using different units in the residual score calculation.}
\label{fig:residual}
\end{figure}

AUC and FPR are calculated using varying numbers of residual units, ranging from 0 to 384, corresponding to the highest to lowest principal units, as illustrated in Fig. \ref{fig:residual}. The official implementation does not specify the number of residual units to use when the dimensionality of the representation is lower than 512. To maintain consistency, we follow the same ratio as used in ViT (starting from unit 512 to the last 768 units). Since the dimensionality in Pathology-SSL is 384, we keep the same \(1/3\) ratio and calculate Residual score starting from unit 256. However, the best OOD detection performance is observed from around unit 23, which use almost the whole principle space in OOD score calculation.

\end{document}